\documentclass[sigconf]{acmart}
\usepackage{subfigure}
\usepackage{multirow}
\usepackage{amsmath} 
\usepackage{hyperref}
\usepackage{balance}
\usepackage[font=small]{caption} 
\AtBeginDocument{%
  }

\setcopyright{acmlicensed}
\copyrightyear{2023}
\acmYear{2023}

\acmConference[MM '23]{Proceedings of the 31st ACM International Conference on Multimedia}{October 29-November 3, 2023}{Ottawa, ON, Canada}
\acmBooktitle{Proceedings of the 31st ACM International Conference on Multimedia (MM '23), October 29-November 3, 2023, Ottawa, ON, Canada}
\acmPrice{15.00}
\acmISBN{979-8-4007-0108-5/23/10}
\acmDOI{10.1145/3581783.3612085}

\acmSubmissionID{1736}

\begin{document}

\title[Incorporating Knowledge Graph
Embedding into Movie Genre Classification]{Incorporating Domain Knowledge Graph into Multimodal Movie Genre Classification with
Self-Supervised Attention and Contrastive Learning}

\author{Jiaqi Li}

\affiliation{%
  \institution{Southeast University}
  \city{Nanjing}
  \country{China}        
}
\email{jqli@seu.edu.cn}

\author{Guilin Qi}
\authornote{Corresponding authors}
\affiliation{%
  \institution{Southeast University}
  \city{Nanjing}
  \country{China}
}
\email{gqi@seu.edu.cn}

\author{Chuanyi Zhang}
\affiliation{%
  \institution{Hohai University}
  \city{Nanjing}
  \country{China}
}
\email{20231104@hhu.edu.cn}

\author{Yongrui Chen}
\affiliation{%
  \institution{Southeast University}
  \city{Nanjing}
  \country{China}
}
\email{yrchen@seu.edu.cn}

\author{Yiming Tan}
\affiliation{%
  \institution{Southeast University}
  \city{Nanjing}
  \country{China}
}
\email{230189757@seu.edu.cn}

\author{Chenlong Xia}
\affiliation{%
  \institution{Southeast University}
  \city{Nanjing}
  \country{China}
}
\email{213203677@seu.edu.cn}

\author{Ye Tian}
\affiliation{%
  \institution{Southeast University}
  \city{Nanjing}
  \country{China}
}
\email{220224335@seu.edu.cn}

\renewcommand{\shortauthors}{Jiaqi Li et al.}

\newcommand{\blue}[1]{{\color{blue}{#1}}}

\begin{abstract}

 Multimodal movie genre classification has always been regarded as a demanding multi-label classification task due to the diversity of multimodal data such as posters, plot summaries, trailers and metadata.  Although existing works have made great progress in modeling and combining each modality, they still face three issues: 1) unutilized group relations in metadata, 2) unreliable attention allocation, and 3) indiscriminative fused features. Given that the knowledge graph has been proven to contain rich information, we present a novel framework that exploits the knowledge graph from various perspectives to address the above problems. As a preparation, the metadata is processed into a domain knowledge graph. A translate model for knowledge graph embedding is adopted to capture the relations between entities. Firstly we retrieve the relevant embedding from the knowledge graph by utilizing group relations in metadata and then integrate it with other modalities. Next, we introduce an Attention Teacher module for reliable attention allocation based on self-supervised learning. It learns the distribution of the knowledge graph and produces rational attention weights. Finally, a Genre-Centroid Anchored Contrastive Learning module is proposed to strengthen the discriminative ability of fused features. The embedding space of anchors is initialized from the genre entities in the knowledge graph. To verify the effectiveness of our framework, we collect a larger and more challenging dataset named MM-IMDb 2.0 compared with the MM-IMDb dataset. The experimental results on two datasets demonstrate that our model is superior to the state-of-the-art methods. Our code and dataset is available at \href{https://github.com/aoluming/IDKG.git}{IDKG.git}.
\end{abstract}




\maketitle

\section{Introduction}
\label{section:intro}

Movie genre classification is a fundamental task for certain downstream tasks such as movie recommendation \cite{choi2012movie}, understanding \cite{islam2022long}, editing \cite{bruckert2022look}, description \cite{rohrbach2017movie}, etc. Previous studies \cite{ertugrul2018movie, yadav2020unified} have achieved unparalleled results in movie genre classification {with} a single modality such as posters, plot summaries, movie trailers, audio, or metadata. Nowadays more researchers \cite{cascante2019moviescope,huang2020movienet,bain2020condensed,zhang2022effectively,vielzeuf2018centralnet,kiela2019supervised,braz2021image,sankaran2021refining} focus on multimodal sources which could be the arbitrary combination of multiple {modalities}. {By taking advantage of multimodal information, existing methods have made great progress in movie genre classification. However, they still leave three issues unsolved}:


\begin{figure}[th]
\centering  
\includegraphics[width=0.8\linewidth]{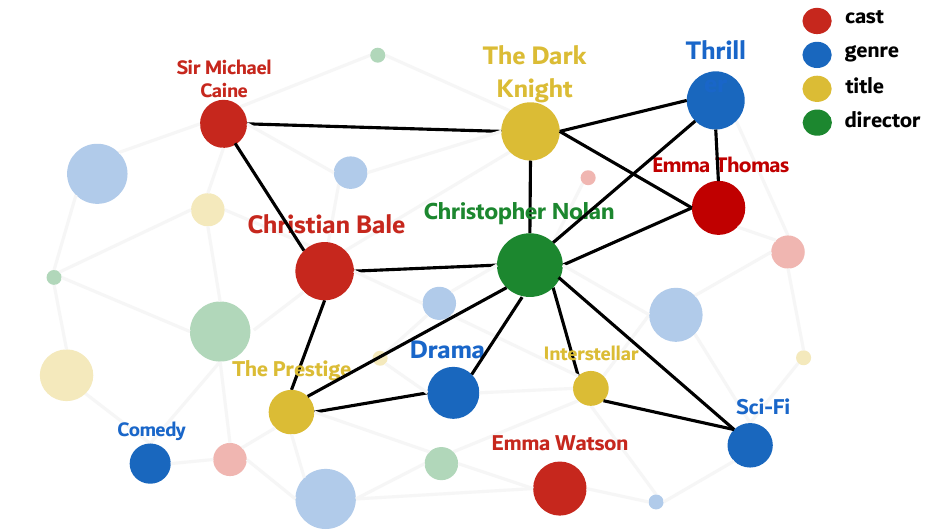}    
\caption{Group relations in metadata. A domain knowledge graph is constructed using titles, casts, directors and genres. An edge between two entities represents their co-appearance in a movie.
}
\label{figintro}
\end{figure}

\label{sec:in}1) Unutilized group relations in metadata. As illustrated in Figure \ref{figintro}, group relations indicate that entities belonging to the same group usually appear simultaneously. To give two real-scenario examples, if \textit{Nolan} is the director of a movie, it is likely to be a \textit{science fiction}. If \textit{Emma Watson} starred in a movie, it is probably not a \textit{comedy}. However, recent works typically ignore group relations when modeling metadata. Behrouzi et al. \cite{behrouzi2022multimodal} extracts features from metadata and fuses them with other modality features through random forest classifier. Seo et al. \cite{seo2022mm} trains a graph attention network based on an undirected graph composed of movie nodes. In our opinion, these methods can be further improved if taking group relations into consideration.


2) Unreliable attention allocation. {Intuitively, different samples should own varying weights on each modality to boost movie genre prediction with multimodal data.}
{Previous works}\cite{arevalo2017gated,behrouzi2022multimodal} {adopt} an attention module to assign different weights to {various} modalities (e.g., plot summary, poster, trailer, audio). {Nevertheless, reliability is not guaranteed due to no supervision in attention module training. Consequently, the produced attention can be irrational.}

3) Indiscriminative fused features.
{Existing methods \cite{arevalo2017gated,yu2022coca,vielzeuf2018centralnet,braz2021image,sankaran2021refining,xu2023bridge} typically utilize a pre-trained model for each modality to obtain discriminative single-modality features. Then features from each modality are fused for genre classification. However, fused feature space tends to show some distance from the original feature space of each modality, which harms the discriminative ability. As a result, fused features tend to be inefficient in predicting genres.}




{Inspired by previous methods \cite{yao2017incorporating,bevilacqua2020breaking,wang2022multi} that leverage knowledge graph, we propose a novel framework named IDKG (\textbf{I}ncorporating \textbf{D}omain \textbf{K}nowledge \textbf{G}raph) for movie genre classification. Our motivation is to exploit the knowledge graph {from} different perspectives to solve the aforementioned issues.}
To begin with, we propose to construct a domain knowledge graph using metadata which includes directors, casts, titles and genres. Moreover, we adopt a translate model for knowledge graph embedding (such as TransH \cite{wang2014knowledge}, TransR \cite{lin2015learning}, etc.) to capture the relation among entities in knowledge graph.
For the first issue, we leverage group relations present in the metadata to retrieve the pertinent embedding from the knowledge graph. {Then the retrieved embedding is integrated with other modalities to improve classification accuracy.}
To alleviate the unreliable attention allocation problem, we propose an Attention Teacher (AT) module that guides the attention module to {produce} rational attention scores based on self-supervised learning. 
{Our AT module captures the distribution feature of the knowledge graph to generate pseudo labels for attention scores and utilizes a suitably designed loss function to train the attention module.}
As to the indiscriminative fused features,
{we propose a Genre-Centroid Anchored Contrastive Learning (G-CACL) module to strengthen the discriminative ability of features. It can be hard to select positive and negative pairs for samples with multiple genres in contrastive learning.
To solve this problem, our G-CACL module defines the centroid of multiple genres embedding as the positive anchor. The enlarged genre space provides feasible optimization directions for fused features to enhance their discriminative ability}.

{In order to verify the effectiveness of our proposed IDKG, we further create a new dataset, MM-IMDb 2.0, which is more challenging compared with MM-IMDb dataset.}
It comprises 33,742 movies collected from IMDb website, with genres set the same as MM-IMDb. Notably, the proportion of the number of head and tail genres is enlarged, thereby enhancing the task difficulty. 

Our main contributions of can be summarized as follows:

\textbullet\ {We propose a novel framework called IDKG that subtly exploits the knowledge graph from different perspectives}. To the best of our knowledge, we are the first to incorporate a knowledge graph into the multimodal movie genre classification task. 

\textbullet\ {We utilize group relations in metadata to obtain relevant embedding from the knowledge graph. With selected embedding as an additional modality source, performance is significantly improved.} 



\textbullet\ {We propose an AT module to alleviate the unreliable attention allocation problem. It obtains pseudo labels from the distribution of the knowledge graph and trains the attention module in a self-supervised manner. Owing to more reliable attention scores, each modality is assigned a more reasonable weight.}


\textbullet\ {We propose a G-CACL module to alleviate the indiscriminative fused features problem. Centroids of genre embedding from the knowledge graph are regarded as positive anchors. The contrastive learning strategy is applied to enhance fused feature representation.}


\textbullet\ We create a new and more challenging dataset MM-IMDb 2.0 to verify the effectiveness of our proposed method. Extensive experiments are conducted to compare our IDKG with current state-of-the-art methods. Experimental results demonstrate that our method outperforms them by a huge margin.

\section{Related Work}

\subsection{Movie Genre Classification}
\label{section:2.1}
Movie genre classification could be divided into two categories: single-modality-based and multimodal-based. The former predicts genres by one kind of modality data, such as posters, plot summaries, movie trailers, audios, or metadata. \cite{simoes2016movie,yadav2020unified} extract features from movie trailers, while \cite{wi2020poster} and \cite{ertugrul2018movie} focus on using only poster images or plot summaries. As single-modality data is relatively simple to process, these methods have achieved excellent performance. However, for multimodal-based methods \cite{behrouzi2022multimodal,ben2018deep,fish2020rethinking} which combine two or more modalities , the results have not been as good due to the complexity of data features.

\subsection{Self-supervised Learning}   
\label{section:2.2}
Self-supervised learning is a special learning method without direct supervision signal. The supervision signal is generated by  using the features of the dataset itself. Nowadays, self-supervised learning is attracting more attention from researchers, such as Next Word Prediction \cite{iter2020pretraining,devlin2018bert}, Automated Text Augmentation \cite{meng2021coco,giorgi2021declutr} in natural language processing and Colorization \cite{vondrick2018tracking,zhang2016colorful}, Context Prediction \cite{noroozi2016unsupervised,misra2016shuffle} in computer vision. Moreover, self-supervised learning \cite{misra2020self,hendrycks2019using,baevski2022data2vec,chen2022scaling} has great potential to replace fully supervised learning in representation learning domain. 

In our paper, we provide a novel insight into the self-supervised learning. We develop a paradigm to the attention module by summarizing the distribution from constructed domain knowledge graph.

\begin{figure*}[t]
\centering
\includegraphics[width=0.99\textwidth]{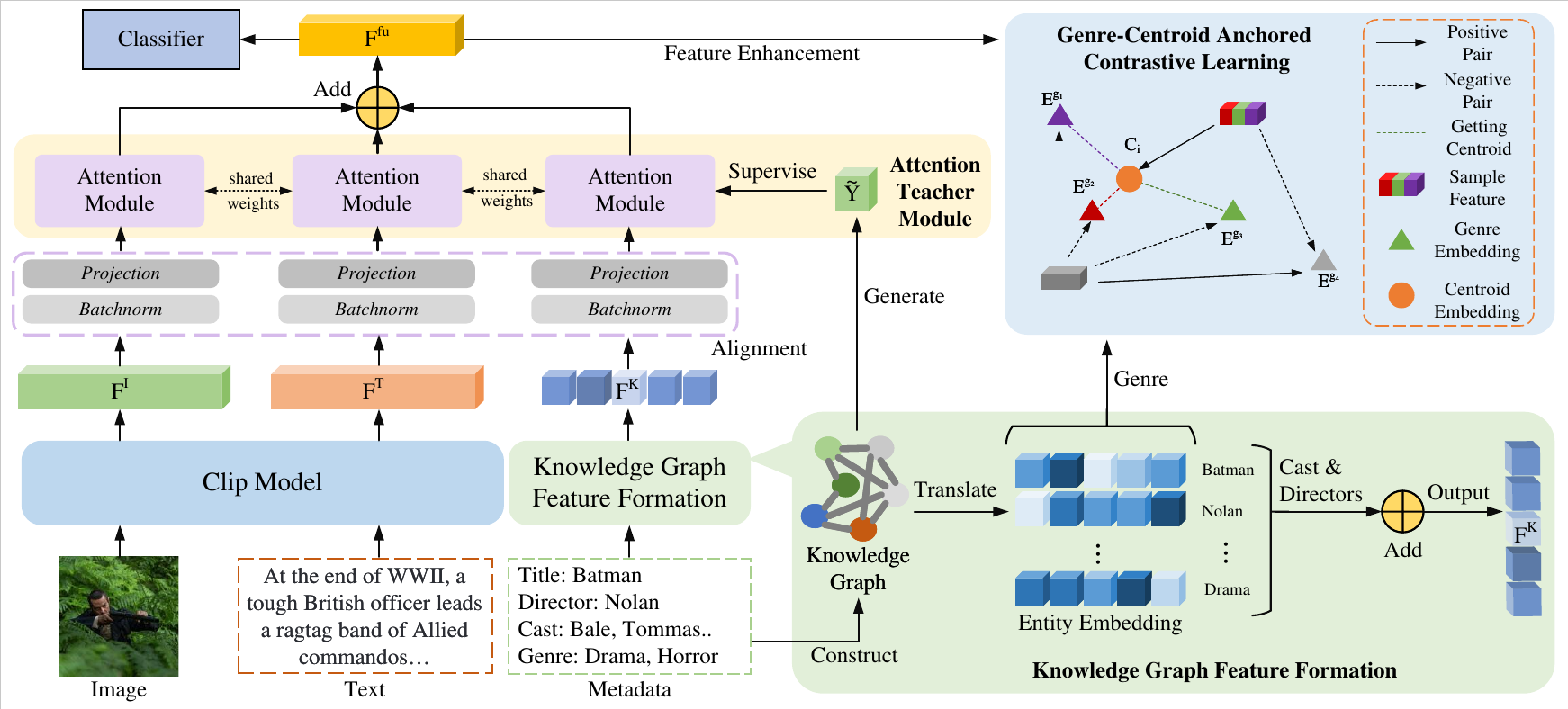}
\caption{The Overview of our proposed IDKG. We leverage a Clip model to extract visual-textual features. Metadata is utilized to construct a domain knowledge graph, which is then followed by a translate model to obtain entity embedding. The cast and director embedding is regarded as the knowledge graph modality feature. Visual, text and knowledge graph features ($F^{I}$, $F^{T}$ and $F^{K}$) are aligned to the same dimension. Next, each modality feature is weighted by the attention module and added to obtain fused feature $F^{fu}$. The Attention Teacher guides the attention producing with pseudo labels $\widetilde{Y}$ generated from the knowledge graph. After that, the discriminative ability of fused feature is enhanced by a Genre-Centroid Anchored Contrastive Learning module, where the embedding space of anchors $ C $ is initialized from the genre embedding $E^{g}$ in the knowledge graph. Finally, the classifier takes the fused feature $F^{fu}$ as input to produce predictions.}
\label{figmodel}
\end{figure*}

\subsection{Supervised Contrastive Learning in Multi-label Classification}
\label{section:2.3}
Contrastive learning is a technique that trains a model to differentiate between similar and dissimilar examples. Such method can be used to learn representations of data. Initially, this approach \cite{park2022fair,wang2022contrastive,han2021dual} was explored in the self-supervised setting. The feature embedding is learned without explicit labels by solving a pretext task. Supervised contrastive learning \cite{khosla2020supervised,gunelsupervised} is another form of contrastive learning that employs annotated data to generate positive pairs by selecting samples from various instances of a specific category. 

 In the contrastive learning paradigm, positive and negative pairs are defined by semantic similarity. Neverthless, it is hard to be applied to multi-label classification because of multiple semantics. Recently, several work attends to bridging combination of multi-label and contrastive learning. \cite{dao2021contrast,wang2022contrastive} compute the contrastive loss by determining a single label that best matches each sample. However, such methods also serve to increase the distance between the sample and other labels. \cite{zhang2022use} proposes a multi-label contrastive learning framework that utilizes a hierarchical structure to leverage all available labels, but in movie genre classification there is no hierarchical structure of labels. \cite{HASSANIN2022103448,pmlr-v162-bai22c} aim to utilize the label embedding space. \cite{HASSANIN2022103448} adopts a center loss but fails to construct negative pairs. \cite{pmlr-v162-bai22c} focuses on figuring out the similarity between label embedding, which may harm the discriminative ability of fused features. To alleviate the problem of defining positive pairs, our proposed G-CACL module enlarges the genre embedding space and a centroid of genres is generated as the positive anchor for each sample. Negative samples are well-designed for the effectiveness of our module.


\section{Approach}

\textbf{Problem Definition.} In multimodal movie genre classification task, the $i$-th sample is composed of a text $T_i$, an image $I_i$ and metadata ${Meta}_i$. ${Meta}_i$ includes the directors, casts, title and the genres of the movie. Each sample is annotated with a multi-label vector $y_i\in \left \{ 0, 1 \right \} ^{M} $, where $M$ is the number of genres of the dataset. The dataset $\mathcal{D} $ is splited into train set $\mathcal{D}_{train}$, test set $\mathcal{D}_{test}$ and validation set $\mathcal{D}_{valid}$. The goal of our task is to train a strong classifier using $\mathcal{D}_{train}$ to predict the multiple genres of each sample in $\mathcal{D}_{test}$.

\textbf{Model Overview.} The overall architecture of our framework is illustrated in Figure \ref{figmodel}. Firstly, a Clip model \cite{radford2021learning} extracts features $F_i^T$ and $F_i^I$ from input source image $I$ and text $T$. Then, in Section \ref{section:3.1}, we process the metadata from $\mathcal{D}_{train}$ into a knowledge graph. To capture the relations between entities, an embedding matrix of knowledge graph entities is trained by adopting a translate model for knowledge graph embedding. We retrieve all the directors and actors in metadata, and get their embedding from embedding matrix as a new modality $K_i$. Next, in Section \ref{section:3.2}, we describe an AT module which could ensure the rational allocation of attention module. Multi-modality features are fused according to their attention weights. Finally, in Section \ref{section:3.3},  we introduce a G-CACL module that enhances the discriminability of fused features.


\subsection{Knowledge Graph Feature Formation}
\label{section:3.1}

\textbf{Knowledge Graph Construction.} A domain knowledge graph is constructed full-automatically by using metadata. Our knowledge graph schema is derived from the fields of metadata and we define four types of entities as the name of fields:
\begin{equation}
\label{for11}
    E=\left \{ d, t, c, g \right \},
\end{equation}
which correspond to directors, titles, casts and genres respectively. Moreover, we define six kinds of relations:
\begin{equation}
\label{for22}
    R=\left \{ d-t, c-t, g-t, d-c, d-g, c-g \right \},
\end{equation}
which represent the relations of directors and titles, casts and titles, genre and titles, directors and cats, directors and genres, casts and genres. Notably we only use metadata from $\mathcal{D}_{train}$ to avoid data leakage. We visit the fields of metadata and extract each value as an entity which has a unique matching id. Furthermore, we traverse all entity pairs in metadata, and each entity pair forms a triplet with the corresponding $R$.


\textbf{Knowledge Graph Embedding.} In order to  capture the relations between entities, we apply a translate model for knowledge graph embedding, for instance, TransH \cite{wang2014knowledge}. Finally we get the embedding matrix of all entities ${Mat}_e\in \mathbb{R}^{N_{k}\times D_{k} } $, where $N_{k}$ denotes the number of entities in knowledge graph and $D_{k}$ is the dimension of entity embedding.


\textbf{Utilization of Group Relations in Metadata.} The group relations in metadata can aid the prediction of movie genres as illustrated in Section \ref{sec:in}. Since the translate model has enabled knowledge graph capture the relations between entities, for $i$-th sample we traverse all the director and cast entities in metadata. Finally we obtain the embedding of entities from ${Mat}_e$: 
\begin{equation}
\label{for33}
    K_i=\left \{ E_i^{d_1},E_i^{d_2},...,E_i^{d_{N_{di}}},E_i^{c_1},E_i^{c_2},...,E_i^{c_{N_{ci}}} \right \},
\end{equation}
where $E^d$ denotes the embedding of an director entity. $N_{di}$ and $N_{ci}$ are the number of directors and casts of the $i$-th sample respectively. Finally, the feature of knowledge graph $F_i^K\in \mathbb{R}^{D_{k} }$ is defined as the sum of all embedding in $K_i$. Notably, in the test phase all $E^d$ and $E^c$ of many samples do not exist in ${Mat}_e$. It is because that ${Mat}_e$ is trained by the entities of $\mathcal{D}_{train}$.
$F^K$ becomes the 0 vector at this point and it can be  formulated as: 

\begin{equation}
\label{for44}
    F_i^{K}=\left\{
\begin{aligned}
\sum_{E^{j}\in K_i}^{}E^j & , & K_i\cap {Mat}_e\ne \emptyset , \\
0 & , & else.
\end{aligned}
\right.
\end{equation}


\subsection{Attention Teacher Module}
\label{section:3.2}



We adopt an attention module to balance the weight of each modality for each sample. It is composed of a linear function and a sigmoid function. Notably the parameters of the attention module for three modality features are shared. 

As illustrated in Figure~\ref{figmodel}, for multi-modality features $F_i^T\in \mathbb{R}^{D_{t} }$, $F_i^I\in \mathbb{R}^{D_{i} }$ and $F_i^K\in \mathbb{R}^{D_{k} }$, where $D_{t}$ and $D_{i}$ are the dimension of text feature and image feature extracted from Clip model, we apply batchnorm1d and linear projection function to convert them to the same shape $D_{p}$. After alignment, these three feature maps are entered into the attention module to obtain their attention scores $A_i^T$, $A_i^I$ and $A_i^K$. Next, these multi-modality features are multiplied by their corresponding attention scores and add up as the ultimate fused feature ${F}_i ^{fu}\in \mathbb{R}^{ D_{p} }$:
\begin{gather}
\label{foratt}
    h(x)=project(batchnorm(x)),\notag\\
     {F}_i ^{fu}=h(F_i^T) \cdot A_i^T + h(F_i^I) \cdot A_i^I+h(F_i^K) \cdot A_i^K,
\end{gather}
where $batchnorm$ and $project$ denotes the batchnorm1d and linear projection function respectively.

To ensure the reliable attention allocation of the attention module, we introduce the Attention Teacher (AT) module based on self-supervised learning. Unlike the previous methods which mainly create pretext tasks in vision or language domain, we mine the distribution feature from the knowledge graph. It is observed that different samples contain different numbers of entities when forming $F^K$. Thus it is natural to consider that varying scores of attention should be assigned to $F^K$ of different samples. Specifically, $F^K$ composed of a small number of entities deserves a lower attention score. Especially in testing phase, if $F^K$ is 0 vector as is presented in Formula \eqref{for44}, the attention scores should be close to 0 at this point. Moreover, we consider the degree of entities in knowledge graph, because an entity is less important if it has very few neighbours in the knowledge graph structure.

Taking above discussions into consideration, we finally utilize the following equations to define the pseudo label $\widetilde{Y}_i$ of attention scores for $F_i^K$:
\begin{gather}
    \overline{N_{d+c}} = \frac{1}{W} \sum\limits_{u=1}^{W}{(N_{du}+N_{cu})}, \notag\\
    \overline{V}=\frac{1}{W} \sum\limits_{u=1}^{W}\sum\limits_{E^{j}\in K_u}^{}V_{E^j}, \notag\\
\widetilde{Y}_i =\frac{(N_{di}+N_{ci})\sum\limits_{E^{j}\in K_i}^{}V_{E^j}}{(N_{di}+N_{ci}+\overline{N_{d+c}})(\sum\limits_{E^{j}\in K_i}^{}V_{E^j}+\overline{V})},
\label{for55}
\end{gather}
where $W$ denotes the number of samples in $\mathcal{D}_{train}$ and $V$ is the degree of an entity in the knowledge graph. In Formula \eqref{for55}, we introduce the average number of sum of directors and actors $\overline{N_{d+c}}$ and the average degree $\overline{V}$ of all samples in  $\mathcal{D}_{train}$. Our intention is that if the sum of directors and actors and the degree of current sample are both beyond the average number in $\mathcal{D}_{train}$, the corresponding pseudo label $\widetilde{Y}_i $ should become higher, otherwise it would be lower. Formula \eqref{for55} requires the range of $\widetilde{Y}_i $ in $(0,1)$ to fit the range of produced attention weights which are limited by a sigmoid function.




Then we design an appropriate loss function to make the self-supervised attention scores converge normally. Since $A_i^K$ is a vector while $\widetilde{Y}_i $ is in scalar form according to Formula \eqref{for55}, $A_i^K$ is averaged as scalar form either to compute loss with $\widetilde{Y}_i $. Since the goal of AT module is to guide $A_i^K$ close to $\widetilde{Y}_i $, a regression loss is adopted in this module. Moreover, it is worth noting that $A_i^K$ and $\widetilde{Y}_i $ are both designed to be between 0 and 1. Thus directly applying l1 or l2 regression loss would limit the loss value in range (0,1) and have the risk of underfitting. In order to ensure a large enough gradient, we apply a logarithmic function to guarantee larges gradients instead of directly adopting l1 or l2 regression loss.
Considering a batch of input with batchsize $B$, the self-supervised attention loss is defined as follows:
\begin{equation}
\label{for66}
    \mathcal{L} _{atten}=-\sum_{i=1}^{B} log(1-|A_i^K-\widetilde{Y}_i |).
\end{equation}
In this way the definition domain of loss function is limited to 0 to 1 with large enough gradient and monotonic increasing trend.
From Formula \eqref{for66}, it can be observed that only the knowledge graph attention score $A_i^K$ is supervised by its pseudo label $\widetilde{Y}_i $ to train the attention module, while $A_i^T$ and $A_i^I$ do not participate in the training procedure. Nevertheless, in our experiment (Section \ref{section:4.4}) we find that the attention module trained with Formula \eqref{for66} can produce reasonable scores $A_i^T$ and $A_i^I$ for texts and images.



\subsection{Genre-Centroid Anchored Contrastive Learning Module}
\label{section:3.3}

We propose a Genre-Centroid Anchored Contrastive Learning (G-CACL) module which facilities the genre embedding from knowledge graph to strengthen the discriminative ability of ${F} ^{fu}$ . Considering that in contrastive learning, each sample is typically assigned a single semantic label and positive pairs are defined by whether they belong to the same semantic label. However, in our task each sample is annotated multi-genre. If we regard each genre embedding as a positive anchor, it is hardly achievable to push the feature of sample close to all the positive anchors synchronously. To overcome this limitation, we attempt to represent the semantics of multiple genres in single anchor. 

Specifically, for a batch of fused features ${F}^{fu}$, the genre embedding set of $i$-th feature ${F}_i ^{fu}$ is: 
\begin{equation}
\label{for77}
   G_i=\left \{ E^{g_1},E^{g_2},...,E^{g_{N_{gi}}} \right \},  
\end{equation}
where $N_{gi}$ is the number of annotated genres of ${F}_i ^{fu}$. we enlarge the genre embedding space by computing the centroid $C_i$ of $G_i$:
\begin{equation}
\label{for88}
   C_i=\frac{1}{N_{gi}} \sum_{E^{g_{k}}\in G_i}^{} E^{g_{k}},  
\end{equation}
which is regarded as the positive anchor for ${F}_i ^{fu}$. 
 We could obtain the union of  genre embedding of all samples in the current batch:
\begin{equation}
\label{for99}
    {\textstyle \bigcup_{i=1}^{B}}G_i = \left \{ E^{g_1},E^{g_2},...,E^{g_{N_{gj}}} \right \},  
\end{equation}
where $N_{gj}$ is the number of annotated genres of all samples in the current batch. We define the complement set of $G_i$ in ${\textstyle \bigcup_{i=1}^{B}}G_i$ as the negative samples of ${F}_i ^{fu}$:
\begin{equation}
\label{for110}
   S_i^{Neg}=  {\textstyle \bigcup_{i=1}^{B}}G_i - G_i.  
\end{equation}
Before computing loss, the centroid and genre embedding go through the linear function to be transformed into the same shape as ${F} ^{fu}$:
\begin{gather}
\label{for111}
    f_{C_i}=Linear(C_i),\notag\\
    f_{E^{g_{k}}}=Linear(E^{g_{k}}).
\end{gather}
The loss function of G-CACL module is defined as follows:
\begin{gather}
\label{for112}
q_i= \sum\limits_{E^{g_k}\in S_i^{Neg}}^{} exp({F}_i ^{fu}\cdot f_{E^{g_k}}/\tau), \notag \\
\mathcal{L}_{contra}=-\sum\limits_{i=1}^{B} log\frac{exp({F}_i ^{fu}\cdot f_{C_i}/\tau)}{exp({F}_i ^{fu}\cdot f_{C_i}/\tau)+q_i},
\end{gather}
where $\tau$ is the temperature coefficient following \cite{khosla2020supervised}. In this loss function, we push the ${F}_i ^{fu}$ close to its corresponding $f_{C_i}$, and away from the embedding of negative samples $S_i^{Neg}$. $q_i$ denotes the sum of similarity between ${F}_i ^{fu}$ and each embedding in $S_i^{Neg}$.


Furthermore, for the multi-label classification, we adopt the binary cross-entropy loss. For a batch of output after the classifier, the multi-label classification loss is:
\begin{equation}
\label{for113}
    \mathcal{L}_{class}=\sum\limits_{i=1}^{B}\sum\limits_{j=1}^{M}(1-y_{ij})\cdot log(1-p_{ij})+y_{ij}\cdot logp_{ij},  
\end{equation}
where $p_{ij}$ denotes the predicted probability that the $i$-th sample belongs to the $j$-th genre.

Finally, we compose $\mathcal{L}_{atten}$, $\mathcal{L}_{contra}$ and $\mathcal{L}_{class}$ as the ultimate training loss for our IDKG:
\begin{equation}
\label{for114}
    \mathcal{L}=\mathcal{L}_{class}+ \mathcal{L}_{atten}+\mathcal{L}_{contra}.
\end{equation}

\begin{table}[t]
\caption{
The comparison of genres distribution of MM-IMDb and MM-IMDb 2.0. The genres are arranged in descending order of quantity from left-top to right-bottom. 
}
\label{tab:dis}
\centering
\scalebox{0.84}{
\begin{tabular}{c p{1cm} p{1.4cm} c p{1cm} p{1.4cm}}
\toprule
\textbf{Genre} &\textbf{MM-IMDb} & \textbf{MM-IMDb 2.0}& \textbf{Genre} & \textbf{MM-IMDb} & \textbf{MM-IMDb 2.0}\\
\cmidrule(r){1-1} \cmidrule(r){2-3} \cmidrule(r){4-4} \cmidrule(r){5-6} 
Drama       & 4188 & 4773 & Fantasy     & 498 & 789 \\
Comedy      & 2609 & 2861 & Music       & 415 & 752 \\
Action      & 1617 & 1950 & History     & 418 & 754 \\
Adventure   & 1588 & 1673 & Western     & 344 & 704 \\
Romance     & 1148 & 1364 & Sci-Fi      & 280 & 687 \\
Crime       & 1081 & 1298 & Musical     & 292 & 603 \\
Horror      & 835 & 1179 & Sport       & 245 & 472 \\
Thriller    & 823 & 1096 & Short       & 211 & 428 \\
Biography   & 584 & 846 & War         & 164 & 417 \\
Animation   & 664 & 874 & Documentary & 139 & 272 \\
Family      & 646 & 817 & File-Noir   & 92 & 73 \\
Mystery     & 591 & 800 & & & \\
\bottomrule
\end{tabular}}
\end{table}

\section{Experiment}

\begin{table*}[t]
\renewcommand{\arraystretch}{0.95}
\caption{
Comparison with the state-of-the-art methods. The evaluation metrics are introduced in Section \ref{section:4.1}}
\centering
\scalebox{0.9}{
\begin{tabular}{llcccccccc}
\toprule
\multirow{2}{*}{Type} & \multirow{2}{*}{Model} 
 & \multicolumn{4}{c}{MM-IMDb} & \multicolumn{4}{c}{MM-IMDb 2.0} \\
 \cmidrule(r){3-6} \cmidrule(r){7-10}
 & & Micro & Macro & Weighted & Samples & Micro & Macro & Weighted & Samples\\
\cmidrule(r){1-2} \cmidrule(r){3-6} \cmidrule(r){7-10}
\multirow{8}{*}{Multimodal}
 & GMU \cite{arevalo2017gated}          & 0.630 & 0.541 & 0.617 & 0.630 & 0.617 & 0.575 & 0.607 & 0.588 \\
 & CentralNet \cite{vielzeuf2018centralnet} & 0.639 & 0.561 & 0.631 & 0.639 & 0.622 & 0.594 & 0.619 & 0.606 \\
 & MMBT \cite{kiela2019supervised}          & 0.669 & 0.618 &   -   &   -   & 0.635 & 0.607 & 0.652 & 0.650 \\
 & MFM \cite{braz2021image}            & 0.675 & 0.616 & 0.675 & 0.673 & 0.656 & 0.608 & 0.664 & 0.671 \\
 & ReFNet \cite{sankaran2021refining}     & 0.680 & 0.587 &   -   &   -   & - & - & - & - \\
 & COCA \cite{yu2022coca}     & 0.677 & 0.626 &   0.668   &   0.681   & 0.659 & 0.623 & 0.649 & 0.670 \\
  & BLIP \cite{li2022blip}     & 0.674 & 0.628 &   0.663   &   0.675   & 0.661 & 0.618 & 0.635 & 0.663 \\
 & BridgeTow \cite{xu2023bridge}     & 0.682 & 0.633 &   0.676   &   0.680   & 0.668 & 0.627 & 0.684 & 0.679 \\
\cmidrule(r){1-2} \cmidrule(r){3-6} \cmidrule(r){7-10}
Graphical
 & MM-GATBT\cite{seo2022mm}                         & 0.685 & 0.645 & 0.683 & 0.686 & 0.674 & 0.632 & 0.697 & 0.685 \\
\cmidrule(r){1-2} \cmidrule(r){3-6} \cmidrule(r){7-10}
\textbf{Graphical+Multimodal} & \textbf{IDKG} & \textbf{0.849} & \textbf{0.832} & \textbf{0.848} & \textbf{0.839} & \textbf{0.828} & \textbf{0.811} & \textbf{0.827} & \textbf{0.807} \\
\bottomrule
\end{tabular}
}
\label{tab:2}
\end{table*}

\subsection{Experiment Setup}
\label{section:4.1}
\textbf{Dataset}. We evaluate IDKG on two datasets, MM-IMDb and MM-IMDb 2.0. MM-IMDb dataset is a multi-label movie genre classification dataset released by \cite{arevalo2017gated} which contains 25959 films. Each sample is composed of a poster, a plot summary and metadata which covers the directors, actors, publication year, etc. Following \cite{arevalo2017gated}, we randomly split the dataset into train set, test set and validation set at the ratio of 0.6, 0.3 
and 0.1.  


To further verify our framework, we create a novel and more challenging dataset, MM-IMDb 2.0. We collect 33742 movies with their posters, plots and metadata from the IMDB website.  As is illustrated in Table \ref{tab:dis}, the ratio of the quantity of \textit{Drama} to the quantity of \textit{Film-Noir} is nearly 65:1. Besides, we also constrain several other tail genres. Moreover, we partition our dataset the same proportion as MM-IMDb dataset. 

\textbf{Domain Knowledge Graph}. As presented in Section \ref{section:3.1}, we process the metadata of $\mathcal{D}_{train}$ into a domain knowledge graph. For MM-IMDb dataset, there are 2231455 triplets and 264271 entities in its domain knowledge graph. Since MM-IMDb 2.0 dataset has much more movies, the number of triplets and entities are 3512803 and 318299 respectively.

\begin{figure}[t]
\centering
\includegraphics[width=0.99
\linewidth]{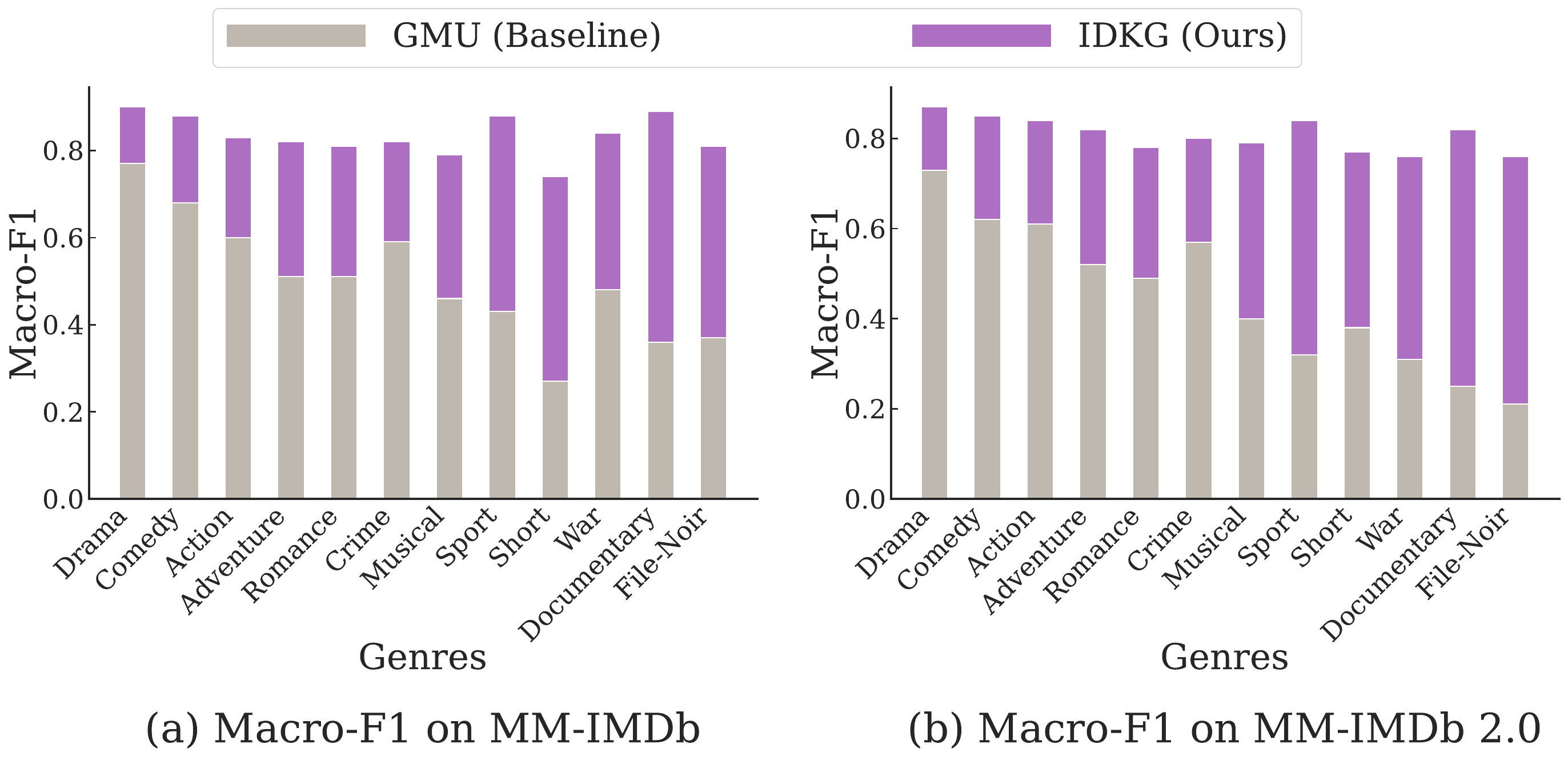}
\caption{Macro-F1 scores for sampled head and tail classes on two datasets. The GMU is chosen as the compared baseline. Genres are arranged in descending order of quantity from left to right.
}
\label{figmoc}
\end{figure}

\textbf{Implementation Details}. IDKG is trained on Pytorch with a 2080ti gpu. For the translate model for knowledge graph embedding, we use the toolkit released by \cite{han-etal-2018-openke}. We adopt the SGD optimizer with 0.5 learning rate and the embedding dimension is 200. Moreover, we train the translate model for 500 epochs with 100 batchsize. For the stage two we use the AdamW \cite{loshchilov2017decoupled} optimizer and the learning rate is 1e-3. We train 15 epochs with 64 batchsize. To extract the image and text features, we apply the Clip \cite{radford2021learning} model and the parameters are frozen in our experiments. The unified vector space dimension is 512 which is set the same as \cite{arevalo2017gated}. The parameter $\tau$ for the G-CACL module is set [0.05, 0.1, 0.3, 0.5, 0.7, 0.9] following \cite{khosla2020supervised}.

\textbf{Evaluation Metrics}. For the evaluation, we report the Micro-F1, Macro-F1, Weighted-F1 and Samples-F1 scores following \cite{arevalo2017gated,seo2022mm} as our metrics.

\subsection{Experimental Results and Analyses}

\label{section:4.2}
We compare IDKG with state-of-the-art methods on two datasets and our approach achieves far superior performance. We categorize the existing methods into three types: 1) \textit{Multimodal} type is a straightforward strategy to solve the problem. This type of works \cite{arevalo2017gated,vielzeuf2018centralnet,kiela2019supervised,braz2021image,sankaran2021refining,xu2023bridge,li2022blip,yu2022coca} mainly focus on the strategy of extracting the well-represented features from each modality respectively by adopting corresponding pre-trained models. It is noted that we also select several recent outstanding \textit{Multimodal} methods \cite{xu2023bridge,li2022blip,yu2022coca} as our competitors.  2) \textit{Graphical} method is conducted to explore the potential of structural sources. MM-GATBT \cite{seo2022mm} leverages graph neural networks to learn the relational semantics of entities by using encoded images as node features. 
3) Our method is a comprehensive framework which fully taking advantage of the two types above.  Notably the translate model used for Table \ref{tab:2}, \ref{tab:3} and \ref{tab:5} is RotateE \cite{sun2019rotate} and the ablation study on translate models is shown in Section \ref{section:4.3}. Moreover, we compare our IDKG with GMU \cite{arevalo2017gated} for head and tail genres on two datasets with Macro-F1 scores as the evaluation metric.

\begin{table*}[t]
\renewcommand{\arraystretch}{1}
\caption{
Ablation study on each module. `\textit{- AT}' represents that AT module is removed and the attention module is not trained with pseudo labels. `\textit{- G-CACL}' means that G-CACL module is omitted and fused features are not enhanced through contrastive learning. `\textit{- KG}' denotes that the metadata is not processed into a knowledge graph, thus not being incorporated with visual-textual features. IDKG (GMU) indicates incorporating the domain knowledge graph (KG) and G-CACL module into GMU.}
\label{tab:3}
\centering
\scalebox{0.95}{
\begin{tabular}{llcccccccc}
\toprule
  \multirow{2}{*}{Model} 
 & \multicolumn{4}{c}{MM-IMDb} & \multicolumn{4}{c}{MM-IMDb 2.0} \\
 \cmidrule(r){2-5} \cmidrule(r){6-9}
  &Micro & Macro & Weighted & Samples & Micro & Macro & Weighted & Samples\\
\cmidrule(r){1-1} \cmidrule(r){2-5} \cmidrule(r){6-9}

   \textbf{IDKG} & \textbf{0.849} & \textbf{0.832} & \textbf{0.848} & \textbf{0.839} & \textbf{0.828} & \textbf{0.811} & \textbf{0.827} & \textbf{0.807} \\
  IDKG - AT & 0.842 & 0.829 & 0.841 & 0.831 & 0.813 & 0.794 & 0.812 & 0.792 \\
  IDKG - AT - G-CACL          & 0.828 & 0.816 &   0.825   &   0.817   & 0.796 & 0.779 & 0.789 & 0.783 \\
  IDKG - AT - G-CACL - KG           & 0.677 & 0.625 & 0.661 & 0.667 & 0.668 & 0.597 & 0.652 & 0.631 \\
  IDKG (GMU) & 0.832 & 0.816 & 0.832 & 0.824 & 0.797 & 0.779 & 0.795 & 0.783 \\
    IDKG (GMU) - G-CACL & 0.819 & 0.804 & 0.817 & 0.810 & 0.782 & 0.773 & 0.790 & 0.772 \\

   IDKG (GMU) - G-CACL - KG                     & 0.630 & 0.541 & 0.617 & 0.630 & 0.617 & 0.575 & 0.607 & 0.588 \\

\bottomrule
\end{tabular}}
\end{table*}

 \textbf{Results on MM-IMDb Dataset}. The comparison results on MM-IMDb dataset are shown in Table \ref{tab:2}. We observe that the Graphical method MM-GATBT \cite{seo2022mm} outperforms all the Multimodal methods in each metric, which demonstrates that the semantic relations in metadata could serve a great benefits to the capacity of predicting genres. Table \ref{tab:2} shows that IDKG surpasses MM-GATBT by at least 15\% on all evaluation metrics. One possible reason may be that in MM-GATBT the graph nodes are composed of the image embedding, where there is no additional knowledge in graph nodes at bottom. IDKG incorporates the knowledge graph embedding with other modalities by using the group relations in metadata which enriches the features for genre prediction. Moreover, two effective modules are designed to address the unreliable attention allocation and indiscriminative fused feature issues, thus boosting the performance of our model. 
 

\textbf{Results on MM-IMDb 2.0 Dataset}. The overall results show that the performance of all methods on MM-IMDb 2.0 dataset is inferior to that on MM-IMDb dataset. The reason can be that MM-IMDb2 2.0 dataset is a more challenging dataset.  As shown in Table \ref{tab:2}, our proposed IDKG also successes in beating all the competitors by at least 12\% which is a large margin on all metrics. The experimental result demonstrates that taking advantage of both Multimodal and Graphical methods can remarkably boost the performance.


\begin{table}[t]
\renewcommand{\arraystretch}{0.9}
\caption{\label{citation-guide}
Ablation study on Translate models. We also report the Hit@10 metrics to illustrate the relation between task performance and the effectiveness of each translate model.
}
\label{tab:4}
\centering
\scalebox{0.84}{
\begin{tabular}{llcccccccc}
\toprule
  Dataset&
  Trans Model 
  &Micro & Macro & Weighted & Samples &Hit@10 \\
\cmidrule(r){1-2} \cmidrule(r){3-6} \cmidrule(r){7-7}  
    \multirow{6}{*}{\parbox{0.9cm}{MM-IMDb}}&TransH \cite{wang2014knowledge} & 0.833 & 0.820 & 0.831 & 0.827 & 0.507\\
  &TransR \cite{lin2015learning} & 0.839 & 0.823 & 0.838 & 0.835 & 0.519\\
  &TransD \cite{ji2015knowledge} & 0.835 & 0.827 & 0.828 & 0.833 & 0.508\\
  &ComplEx \cite{trouillon2016complex} & 0.827 & 0.813 & 0.825 & 0.821 & 0.485\\
  &ConvE \cite{dettmers2018convolutional} & 0.829 & 0.822 & 0.836 & 0.825 & 0.506\\
   &
   \textbf{RotateE} \cite{sun2019rotate} & \textbf{0.849} & \textbf{0.832} & \textbf{0.848} & \textbf{0.839} & \textbf{0.549}\\
\cmidrule(r){1-2} \cmidrule(r){3-6} \cmidrule(r){7-7}  
    \multirow{6}{*}{\parbox{0.9cm}{MM-IMDb 2.0}}&TransH \cite{wang2014knowledge} & 0.813 & 0.806 & 0.815 & 0.792 & 0.507\\
  &TransR \cite{lin2015learning} & 0.814 & 0.809 & 0.820 & 0.791 & 0.519\\
  &TransD \cite{ji2015knowledge} & 0.817 & 0.803 & 0.823 & 0.796 & 0.508\\
  &ComplEx \cite{trouillon2016complex} & 0.806 & 0.797 & 0.812 & 0.784 & 0.485\\
  &ConvE \cite{dettmers2018convolutional} & 0.814 & 0.805 & 0.822 & 0.792 & 0.506\\
   &
   \textbf{RotateE} \cite{sun2019rotate} & \textbf{0.828} & \textbf{0.811} & \textbf{0.827} & \textbf{0.807} & \textbf{0.549}\\
  
\bottomrule
\end{tabular}
}
\end{table}

\textbf{Macro-F1 score analysis}. As shown in Table \ref{tab:dis}, the distribution of genres is imbalanced in MM-IMDb dataset. Towards severer imbalance problem, we enlarge the proportion of the number of head genres and tail genres when collecting MM-IMDb 2.0 dataset as mentioned in Section \ref{section:4.1}. As Macro-F1 neglects the proportion for each label, it is more sensitive to the imbalance of genres distribution than other evaluation metrics. Thus we compare Macro-F1 between GMU and IDKG for sampling six head genres and six tail genres on two datasets as can be seen in Figure \ref{figmoc} and the  genres are arranged in descending order of quantity from left to right. We could observe that for GMU the Macro-F1 of head classes is generally far larger than that of tail classes on two datasets due to the imbalance distribution. However, for IDKG the Macro-F1 of tail genres is close to that of head genres and almost Macro-F1 of all genres is around 80\%, which demonstrates distinguished classification ability of IDKG.



 \subsection{Ablation Study}
\label{section:4.3}

To evaluate the effectiveness of each component of IDKG, we conduct extensive ablation experiments on two datasets.  In particular, we incorporate the domain knowledge graph and the  G-CACL module into GMU \cite{arevalo2017gated} one by one for more solid proof. After the ensemble of all modules, we name the new model IDKG (GMU). Notably since GMU provides a strategy of feature fusion, AT module is not conducted in its ablation experiment. Moreover, we compare the different translate models for knowledge graph embedding and show the influence on IDKG performance. Finally, we analyze the parameter $\tau$ of G-CACL
module.

\begin{figure}[t]
\centering
\includegraphics[width=0.92\linewidth]{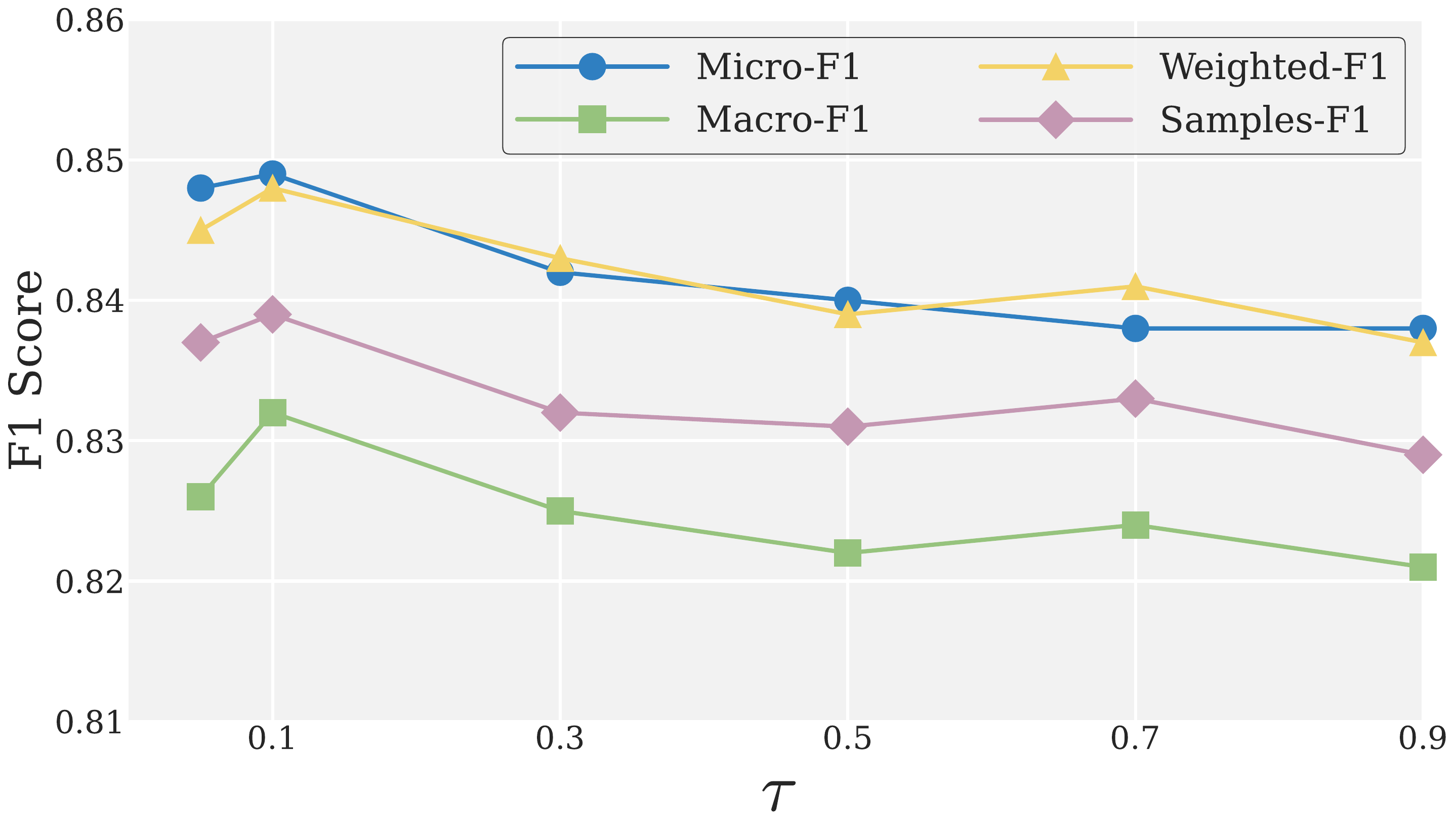}
\caption{The effect of  parameter $\tau$ on MM-IMDb. We report F1 scores on different value of $\tau$.
}
\label{figpara}
\end{figure}

\textbf{Results on IDKG}. Table \ref{tab:3} illustrates that the discarding of domain knowledge graph loses the most performance by at least 10\% of all modules which proves that group relations in metadata could contribute to the model performance greatly. With the removal of AT module, the performance of IDKG on all metrics declines to various degrees on two datasets. This is may be that AT module ensure the reliable attention allocation, thus improving the accuracy. The performance goes down if IDKG is not equipped with G-CACL module. The reason is perhaps that G-CACL module could improve the discriminative ability of fused feature which could boost the effectiveness of our model.

\textbf{Results on IDKG (GMU)}.  As can be seen from Table \ref{tab:3}, the trend of performance change on IDKG (GMU) is the same as IDKG. This illustrates the effectiveness of our proposed modules. It is noted that the performance drops significantly when remove G-CACL module for IDKG (GMU). It demonstrates that GMU is weak in the discriminative ability of fused feature and our G-CACL module compensates the shortcoming.

\textbf{Effect of Translate Models}. We compare different translate models including TransH \cite{wang2014knowledge}, TransR \cite{lin2015learning}, TransD \cite{ji2015knowledge}, ComplEx \cite{trouillon2016complex}, ConvE \cite{dettmers2018convolutional}  and RotateE \cite{sun2019rotate} to observe the effects on IDKG performance. Moreover, we list the Hit@10 results of predicting missing links on WN18 dataset, which are reported in the toolkit \cite{han-etal-2018-openke}. As show in Table \ref{tab:4}, we notice that the RotateE achieves the best performance and ComplEx performs worst. Combining the Hit@10 results, we could draw a conclusion that the performance of IDKG is correspond to the capacity of translate model due to strong embedding representation enhancing the ability of capturing relations between entities.

\textbf{Analysis on parameter $\tau$}. As illustrated in \cite{wang2021understanding}, smaller $\tau$ is sensitive to difficult negative samples that too small $\tau$ would destroy the embedding space due to the unproper negative samples setting. According to Figure \ref{figpara}, when $\tau$ is 0.1 the performance of IDKG is the best on MM-IMDb dataset. The reason may be that in our method the negative pairs are rationally constructed, thus small $\tau$ would benefit the contrastive loss.

\begin{figure}[t]
\centering
\includegraphics[width=\linewidth]{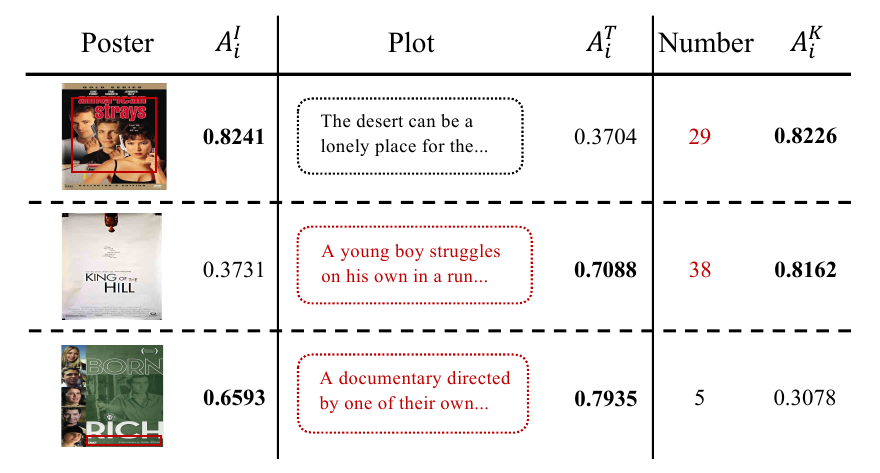}
\caption{Case study on AT module. The modalities marked red relatively merit higher attention weights. We report the corresponding attention score of each modality ($A_i^I$, $A_i^T$ and $A_i^K$) to verify the effectiveness of our AT module.}
\label{figat}
\end{figure}

\subsection{Comparison with Multi-label Contrastive Learning Methods}
\label{section:4.5}
To demonstrate the effectiveness of G-CACL module, we not only conduct extensive ablation study as illustrated in Section \ref{section:4.3}, but also replace it with existing multi-label classification methods which adopt contrastive learning. The competitors are MulCon \cite{dao2021contrast}, MCL \cite{HASSANIN2022103448}, MLTC \cite{wang2022contrastive} and C-GMVAE \cite{pmlr-v162-bai22c} and all of them are introduced in Section \ref{section:2.3}. The results demonstrated in Table \ref{tab:5} verify that our G-CACL module is superior to other contrastive learning methods. We observe that the performance of embedding space initialized randomly is worse than that initialized from the genre embedding of our knowledge graph. We assume that it is because the knowledge graph embedding captures the discrimination between genre semantics, thus improving the effectiveness of our module.

\subsection{Analysis on Attention Teacher Module}
\label{section:4.4}


To test the validity of AT module, we present a case study as shown in Figure \ref{figat}. We record $A_i^I$, $A_i^T$ and $A_i^K$ in testing phase and sample 3 cases, where the modalities that contribute more to the prediction are marked in red. It is observed that the attention module outputs the reliable scores for each modality. Notably despite that $A_i^I$ and $A_i^T$ are not directly supervised, they still output the rational scores. For example, the plot summary of the second and third sample should be allocated more attention weights and $A_i^T$ of them are relative high. We infer that $A_i^I$ and $A_i^T$ are soft-supervised because of the shared parameters of the attention module.

\begin{table}[t]
\caption{\label{citation-guide}
Comparison with other contrastive learning methods on multi-label classification. Notably \textbf{Ours (random)} represents the genre embedding is initialized randomly. 
}
\label{tab:5}
\centering
\scalebox{0.87}{
\begin{tabular}{llcccccccc}
\toprule
  Dataset&
  Trans Model 
  &Micro & Macro & Weighted & Samples \\
\cmidrule(r){1-2} \cmidrule(r){3-6}   
    \multirow{6}{*}{MM-IMDb}&MulCon \cite{dao2021contrast} & 0.830 & 0.818 & 0.827 & 0.816\\
  &MCL \cite{HASSANIN2022103448} & 0.832 & 0.813 & 0.829 & 0.822 \\
  &MLTC \cite{wang2022contrastive} & 0.835 & 0.821 & 0.825 & 0.826 \\
  &C-GMVAE \cite{pmlr-v162-bai22c} & 0.846 & 0.828 & 0.846 & \textbf{0.840} \\
  \cmidrule(r){2-2} \cmidrule(r){3-6}
  &Ours (random) & 0.841 & 0.828 & 0.842 & 0.835 \\
   &
   \textbf{Ours}  & \textbf{0.849} & \textbf{0.832} & \textbf{0.848} & 0.839 \\
\cmidrule(r){1-2} \cmidrule(r){3-6}   
    \multirow{6}{*}{MM-IMDb 2.0}&MulCon \cite{dao2021contrast} & 0.806 & 0.785 & 0.792 & 0.784\\
  &MCL \cite{HASSANIN2022103448} & 0.811 & 0.794 & 0.802 & 0.787 \\
  &MLTC \cite{wang2022contrastive} & 0.815 & 0.799 & 0.812 & 0.792 \\
  &C-GMVAE \cite{pmlr-v162-bai22c} & 0.825 & 0.807 & 0.824 & 0.806 \\
  \cmidrule(r){2-2} \cmidrule(r){3-6}
  &Ours (random) & 0.821 & 0.808 & 0.821 & 0.802 \\
   &
   \textbf{Ours}  & \textbf{0.828} & \textbf{0.811} & \textbf{0.827} & \textbf{0.807} \\
  

\bottomrule
\end{tabular}
}
\end{table}

\section{Conclusion}
In this paper, we proposed an effective and novel framework named IDKG. To the best of our knowledge, we are the first to apply knowledge graph technology to multimodal movie genre classification. Firstly, IDKG utilized the group relations in the knowledge graph to obtain embedding and incorporated it with other modalities. Furthermore, an Attention Teacher module was proposed to learn the distribution of the knowledge graph and guide the attention module to allocate more reliable weights. Finally, a Genre-Centroid Anchored Contrastive Learning module enhanced the discriminative ability of the fused feature. We also collected a new large-scale dataset named MM-IMDb 2.0 for movie genre classification which faces a severer class-imbalanced problem compared with the MM-IMDb dataset. Finally, we conducted extensive experiments on MM-IMDb and MM-IMDb 2.0 datasets and the experimental results demonstrated that the performance of our model was superior to existing methods. For future work, we plan to construct a multimodal domain knowledge graph of the movie field, as well as applying to more downstream tasks.

\begin{acks}
We thank the Big Data Computing Center of Southeast University for providing the facility support on the numerical calculations in this paper.
\end{acks}

\bibliographystyle{ACM-Reference-Format}
\balance
\bibliography{samples/custom}










\end{document}